\documentclass[runningheads]{llncs}
\usepackage{graphicx}

\usepackage[dvipsnames]{xcolor}
\usepackage{enumitem}
\setitemize{noitemsep,topsep=0pt,parsep=0pt,partopsep=0pt}

\begin{document}
\title{Visual FUDGE: \\Form Understanding via Dynamic Graph Editing}

\titlerunning{Visual FUDGE: Form Understanding via Dynamic Graph Editing}
%
\author{Brian Davis\inst{1}* \and
Bryan Morse\inst{1} \and
Brian~Price\inst{2}  \and \\
Chris Tensmeyer\inst{2} 
\and~Curtis Wiginton\inst{2}}
\authorrunning{B. Davis et al.}
%
\institute{Brigham Young University\\
\email{\{briandavis,morse\}@byu.edu}\\\and
Adobe Research\\
\email{\{bprice,tensmeye,wigingto\}@adobe.com}\\
{*} Corresponding author}
\maketitle              
\begin{abstract}


We address the problem of \textit{form understanding}: finding text entities and the relationships/links between them in form images. 
The proposed FUDGE model formulates this problem 
on a graph of text elements (the vertices) and uses a Graph Convolutional Network to predict changes to the graph.
The initial vertices are detected text lines and do not necessarily correspond to the final text entities, which can span multiple lines. Also, initial edges contain many false-positive relationships. FUDGE edits the graph structure by combining text segments (graph vertices) and pruning edges in an iterative fashion to obtain the final text entities and relationships.
While recent work in this area has focused on leveraging large-scale pre-trained Language Models (LM), FUDGE achieves almost the same level of entity linking performance on the FUNSD dataset by learning only visual features from the (small) provided training set. 
FUDGE can be applied on forms where text recognition is difficult (e.g. degraded or historical forms) and on forms in resource-poor languages where pre-training such LMs is challenging. FUDGE is state-of-the-art on the historical NAF dataset.

\end{abstract}
\section{Introduction}

Paper forms are a convenient way to collect and organize information, and it is often advantageous to digitize such information for fast retrieval and processing.
While OCR and handwriting recognition (HWR) methods can extract raw text from a form image, we aim to understand the layout and relationships among the text elements (e.g., that ``Name:'' is associated with ``\textit{Lily Johnson}'').
The term \textit{form understanding} was recently coined~\cite{funsd} as the task of extracting the full structure of information from a form image.
We define the task as: given a form image, identify the semantic text entities and the relationships between them \textit{with no prior form template}. 
This work focuses on finding relationships among text entities, although this requires first detecting, segmenting, and classifying text into entities.
After entities are predicted, the relationship detection or entity linking task is simply predicting which text entities have a semantic relationship. 

Most recent works on form understanding rely primarily on large-scale pre-trained Language Models (LMs)~\cite{layoutlm,wang-etal-2020-docstruct,bros}, which in turn have a dependency on accurate text detection and recognition.
Such approaches may perform poorly in domains where OCR/HWR results are poor or on languages with limited resources to train such LMs. (LayoutLM~\cite{layoutlm}  is pre-trained on 11 million documents.)
OCR/HWR often struggle on damaged or degraded historical documents and on document images captured inexpertly with a smartphone.

In contrast, we present a purely visual solution, improving on our previous visual form understanding method in Davis et al.~\cite{previous}. 
Given forms in an unfamiliar language, humans can generally infer the text entities and their relationships using layout cues and prior experience, which we aim to approximate. 
Our approach doesn't require language information and could be applied to visually similar languages (e.g., those sharing the same script), possibly without fine-tuning.
In this work we attempt to show that a visual model trained on a small dataset without language information is, on several tasks, comparable to methods that rely on large amounts of pre-trained language information.

Similar to some prior works~\cite{previous,wang-etal-2020-docstruct,carbonell}, we model forms as a graph, where text segments are the vertices and pairwise text relationships are edges.
In~\cite{previous} we scored pairwise heuristic relationship proposals independently using visual features and applied global optimization as a post-processing step to find a globally coherent set of edges. We improve upon this by making our model more end-to-end trainable and by jointly predicting relationships with  Graph Convolutional Network (GCN).
Additionally,~\cite{previous} was unable to predict text entities that span multiple text lines, a problem solved with our dynamic graph editing.
An alternative formulation to solve form understanding visually is to treat it as a pixel labeling problem, as in Sarkar et al.~\cite{sarkar2020}.
However, it is not clear from~\cite{sarkar2020} how to infer form structure (bounding boxes and relationships) from pixel predictions, and the proposed (even dilated) CNN model could have difficulty modeling relationships between spatially distant elements.
Instead we use a GCN that directly predicts the form structure and does not need to rely on limited receptive fields to propagate information spatially.


\begin{figure}[t]
\centering
\includegraphics[width=0.99\textwidth]{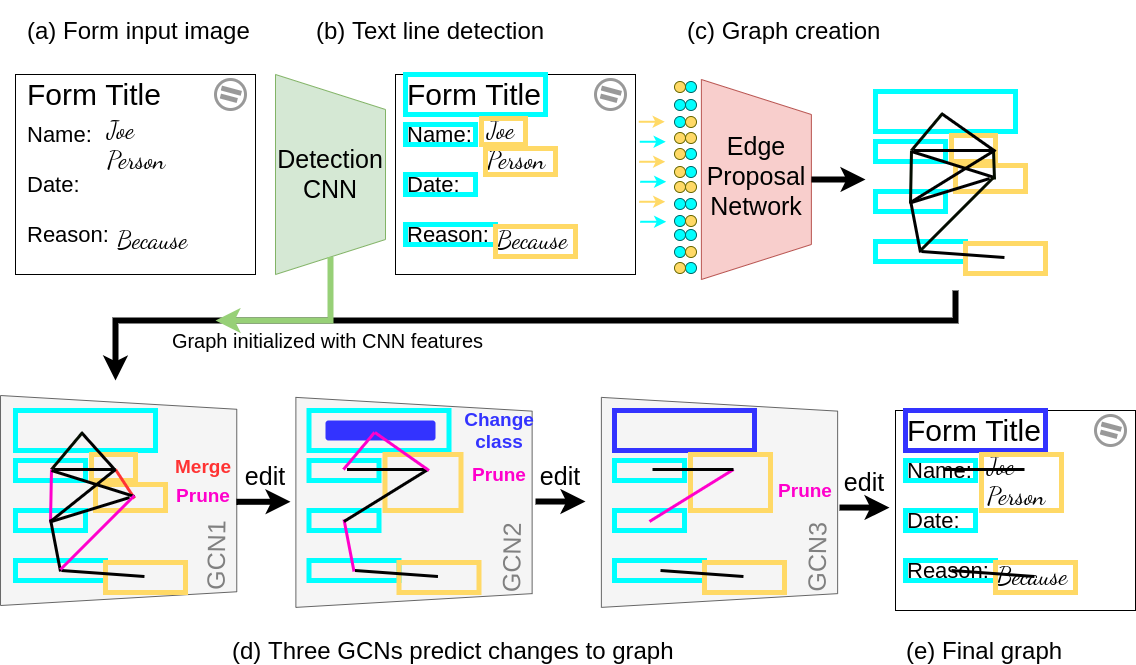}
\label{fig:overview}
\caption{Overview of our model, FUDGE. From a form image (a)  text line detection (b) is performed. Then (c) an edge proposal score is computed for each possible edge. After thresholding the scores, the remaining edges form the graph. The graph is initialized with spatial features and features from the CNN detector. A series of GCNs are run (d), each predicting edits to the graph (pruning irrelevant edges, grouping text lines into single entities, correcting oversegmented lines). The final graph (e) is the text entities and their relationships.} 
\end{figure}

Our proposed FUDGE (Form Understanding via Dynamic Graph Editing) model is a multi-step process involving text line detection, relationship proposals, graph editing to group text lines into coherent text entities and prune edges, leading to relationship prediction (Fig.~\ref{fig:overview}). 
We use GCNs so that semantic and relationship predictions can be jointly predicted.
We initialize the graph vertices with detected text lines (visual  detection of semantically grouped words is hard).
However, the relationships of interest are between text entities, which can be  multiple text lines.
FUDGE is unique from other GCNs as we dynamically edit the graph during inference to predict which vertices should be merged into a single vertex.
This groups text lines into text entities and corrects over-segmented initial text detections.

In summary, our primary contributions are:
\begin{itemize}
    \item a GCN that jointly models form elements in the context of their neighbors,
    \item an iterative method to allow the GCN to merge vertices and prune edges, 
    \item extensive experiments on the  FUNDS~\cite{funsd} and NAF~\cite{previous} datasets which validate that FUDGE performs comparably to methods that pre-train on millions of images.
\end{itemize}
Our code is available at \url{https://github.com/herobd/FUDGE}.

\section{Prior Work}

Automated form processing has been of interest for decades, but generally template-based solutions have been used to extract information from a few known form layouts. Recently, the idea of form understanding~\cite{funsd} has become an area of interest. This has evolved from extraction of information in a template-free setting~\cite{ACP} to capturing all the information a form contains. 
Recent methods have leveraged the astounding progress of large pre-trained language models~\cite{bert}, focusing on language rather than visual elements to understand forms.

DocStruct~\cite{wang-etal-2020-docstruct} is a language-focused approach for form relationship detection. 
They blend three feature sources: semantic/language from BERT~\cite{bert}, spatial text bounding box coordinates, and visual infromation extracted from pixels. 
Their ablation results~\cite{wang-etal-2020-docstruct} show that the language and spatial features are the most valuable, which is reasonable given that
the visual features are extracted only from the text areas and thus ignore most of the visual context around the relationships, especially distant ones. 



BROS~\cite{bros} is an unpublished method that builds on the ideas of LayoutLM~\cite{layoutlm}, which is primarily a BERT model~\cite{bert} with additional spatial and visual features appended to the word embeddings. 
Similar to DocStruct, LayoutLM only extracts visual features immediately around the text instances. 
BROS~\cite{bros} adds better spatial representations and awareness to the model in addition to better pre-training tasks, but do not improve its use of visual features compared to LayoutLM. 
BROS solves both the text entity extraction (detection) task and entity linking (relationship detection). 
Entity extraction is done in two steps: start-of-entity detection and semantic classification, and next-token classification which collects the other tokens composing the entity. 
Relationship detection is done by computing an embedding for each start-of-entity token and doing a dot product multiplication across all pairs of start-of-entity token embeddings, giving the relationship scores.
Form2Seq~\cite{form2seq} is similar to LayoutLM, but uses LSTMs instead of Transformers and omits visual features entirely.

The unpublished LayoutLMv2~\cite{layoutlmv2} is a Transformer based model that uses a much stronger visual component, adding visual tokens to the input and vision-centric pre-training tasks. 
For forms, LayoutLMv2 has only been evaluated on the entity detection and not on relationship detection.
Another example of a good blend of visual and language information is used by both Attend, Copy, Parse~\cite{ACP} and Yang et al.~\cite{Yang_2017_CVPR}. 
Both encode the text as dense vectors and append these vectors to the document image pixels at the corresponding spatial locations.
A CNN can then perform the final task, which is information extraction in~\cite{ACP} and semantic segmentation in~\cite{Yang_2017_CVPR}. 

In contrast, Davis et al.~\cite{previous} and Sarkar et al.~\cite{sarkar2020} propose language-agnostic models.
In~\cite{previous} we focused on the relationship detection problem and used a CNN backbone to detect text lines.
Relationship candidates were found using a line-of-sight heuristic and each candidate was scored using a visual-spatial classifier.
However, each candidate was scored independently, and a separate post-processing optimization step was needed to resolve global consistency issues.
This latter step also required predicting the number-of-neighbors for each text line, which may not be accurate on more difficult datasets.
Sarkar et al.~\cite{sarkar2020} focus on extracting the structure of forms but treat it as a semantic segmentation (pixel labeling) problem.  They use a U-Net architecture, and at the lowest resolution include dilated convolutions to allow information to transfer long distances.
Sarkar et al.\cite{sarkar2020} predicts all levels of the document hierarchy in parallel, making it quite efficient.

Aggarwal et al.~\cite{Aggarwal_2020_WACV} offers an approach that is architecturally like a language-based approach but uses contextual pooling of CNN features like~\cite{previous}. 
They determine a context window by identifying a neighborhood of form elements and use a CNN to extract image features from this context window. 
They also extract language features (using a non-pre-trained LSTM), which are combined with the visual features for the final decisions. 
However, the neighborhood is found using $k$-nearest neighbors with a distance metric which could be sensitive to cluttered or sparse forms and long distance relationships.

GCNs have been applied to other structured document tasks, such as table extraction~\cite{Qasim2019,Riba2019}.
Carbonell et al.~\cite{carbonell} use a GCN to solve the entity grouping, labeling and linking tasks on forms. 
They use word embedding and bounding box information as the initial node features and they do not include any visual features. They use k-nearest neighbors to establish edges. In our method, we update our graph for further GCN processing, particularly grouping entities together. Carbonell et al.~\cite{carbonell} predict the entity grouping from a GCN and then sum the features of the resulting groups. Rather than processing more with a GCN, they predict entity class and linking from these features.
Unlike most other non-visual methods, Carbonell et al.~\cite{carbonell} do not use any pre-training. 





\section{FUDGE Architecture}

FUDGE is based on Davis et al.~\cite{previous}. We use the same detection CNN backbone and likewise propose relationships, extract local features around each proposed relationship, and then classify the relationships. FUDGE differs in three important ways:
\begin{itemize}
    \item In~\cite{previous} we used line-of-sight to propose relationships, which 
    can cause errors due to false positive detections and form layouts that don't conform to the line-of-sight assumption. FUDGE instead learns an edge proposal. 
    \item Instead of predicting each relationship in isolation, we put the features into a graph convolutional network (GCN) so that a joint decision can be made. This also allows semantic labels for the text entities to be predicted jointly with the relationships, as they are very related tasks. 
    \item We allow several iterative edits to the graph, which are predicted by the GCN. 
    Text lines are grouped into single text entities, and oversegmented text lines are corrected, by aggregating groups of nodes into new single nodes. 
    Spurious edges are pruned.
\end{itemize}
Fig.~\ref{fig:overview} shows an overview of FUDGE. We now go into each component in detail.

\subsection{Text line detection}

We use the same text line detector as \cite{previous}, only ours does not predict the number of neighbors. This detector is a fully-convolutional network with wide horizontally strided convolutions and a YOLO predictor head~\cite{yolov3}. We threshold predictions at 0.5 confidence. The detector is both pre-trained 
and fine-tuned during training of the relationship detection. The detection makes an auxiliary class prediction; the final text entity class prediction is made by the GCN. 

\subsection{Edge proposal}

While line-of-sight is effective for the simplified NAF dataset contributed by Davis et al.~\cite{previous}, it is brittle and doesn't apply to all cases. FUDGE instead learns an edge proposal network using a simple two-layer linear network with ReLU activation in between. It receives features from each possible pair of detected text lines and then predicts the likelihood of an edge (they are either oversegmented, part of the same entity, or parts of entities that have a relationship). Half of the relationships with the highest scores (maximum of 900) are used to build the initial undirected graph.
The features are: difference of $x$ and $y$ position for all corresponding corners and the center of the boxes, height and width of both boxes, L2 distance of all corresponding corners, normalized $x$ and $y$ position for both bounding boxes, whether there is a line of sight between the boxes (computed as in~\cite{previous}), and the detection confidences and class predictions for both boxes. We predict both permutations of the pair orders and average them.

\subsection{Feature extraction}

While the relationship proposal step gives us the initial graph structure, we also need to initialize the graph with features. We use a GCN architecture with features on both the nodes and the edges (described in Sec.~\ref{sec:iterate}). We use two types of features: spatial features, similar to those used in edge proposal, and visual features from two layers of the detection CNN (high- and low-level features). We perform an ROIAlign~\cite{roialign} for a context window, and then a small CNN processes those features, eventually pooling to a single vector. This is almost identical to the features extraction in Davis et al.~\cite{previous}, only differing in resolution, padding, and the CNN hyper-parameters.

For nodes, the context window is the bounding box surrounding the text line(s) composing the entity, padded by 20 pixels (image space) on all sides. The ROIAlign pools the features to 10 $\times$ 10 resolution and the two feature layers are appended. Two mask layers are appended to these features: one of all detected text boxes, and the other of just the text boxes belonging to this entity (these are from the same window as the ROIAlign). These are passed to a small CNN which ends with global pooling  (exact network in Table~\ref{tab:feature_ex}).
The bounding box surrounding all the text lines of the entity is used to compute additional spatial features: detection confidence, normalized height, normalized width, and class prediction; these are appended to the global pooled features from the CNN.

For edges, the context window encompasses all the text lines composing the two nodes, padded by 20 pixels. The ROIAlign pools the features into 16 $\times$ 16 resolution (larger than the resolution than for nodes, as more detail exists in these windows). Appended to the CNN features are three mask layers: all detected text boxes, and one for each of the nodes containing all of the text boxes for the entity. These are passed to a small CNN which ends with global pooling (exact network in Table~\ref{tab:feature_ex}).
The two bounding boxes surrounding all the text lines of each entity are used to compute the spatial features: normalized height of both entities, normalized width of both boxes, the class predictions of each entity, and distance between the corner points of the two entities (top-left to top-left, etc.). These are appended to the features from the CNN.

These node and edge features are passed through a single linear node or edge transition layer to form the initial features of the graph.

We ROIAlign high- and low-level features from the CNN (second-to-last conv layer and first pool layer) as high-level features generally contain the more interesting and descriptive information but may leave out certain low-level features that were irrelevant for the detection task.

\begin{table}[tb]
\centering
\caption{Architecture of feature extraction CNNs for nodes and edges.}\label{tab:feature_ex}
\begin{tabular}{|l|l|}
\hline
\textbf{Node} (starts at 10x10, 320 channels)             & \textbf{Edge} (starts at 16x16, 320 channels)             \\
\hline
Depth-wise seperable 3x3 conv, 64  & Depth-wise seperable 3x3 conv, 128 \\
Depth-wise seperable 3x3 conv, 64  & Depth-wise seperable 3x3 conv, 128 \\
Max pool 2x2                      & Max pool 2x2                      \\
Depth-wise seperable 3x3 conv, 128 & Depth-wise seperable 3x3 conv, 256 \\
3x3 conv, 256                       & Depth-wise seperable 3x3 conv, 256 \\
Global average pooling             & Max pool, 2x2                      \\
                                   & 3x3 conv, 256                       \\
                                   & Global average pooling            \\
\hline
\end{tabular}
\end{table}

\subsection{Iterated graph editing with GCN}
\label{sec:iterate}

\begin{figure}[t]
\centering
\includegraphics[width=0.999\textwidth]{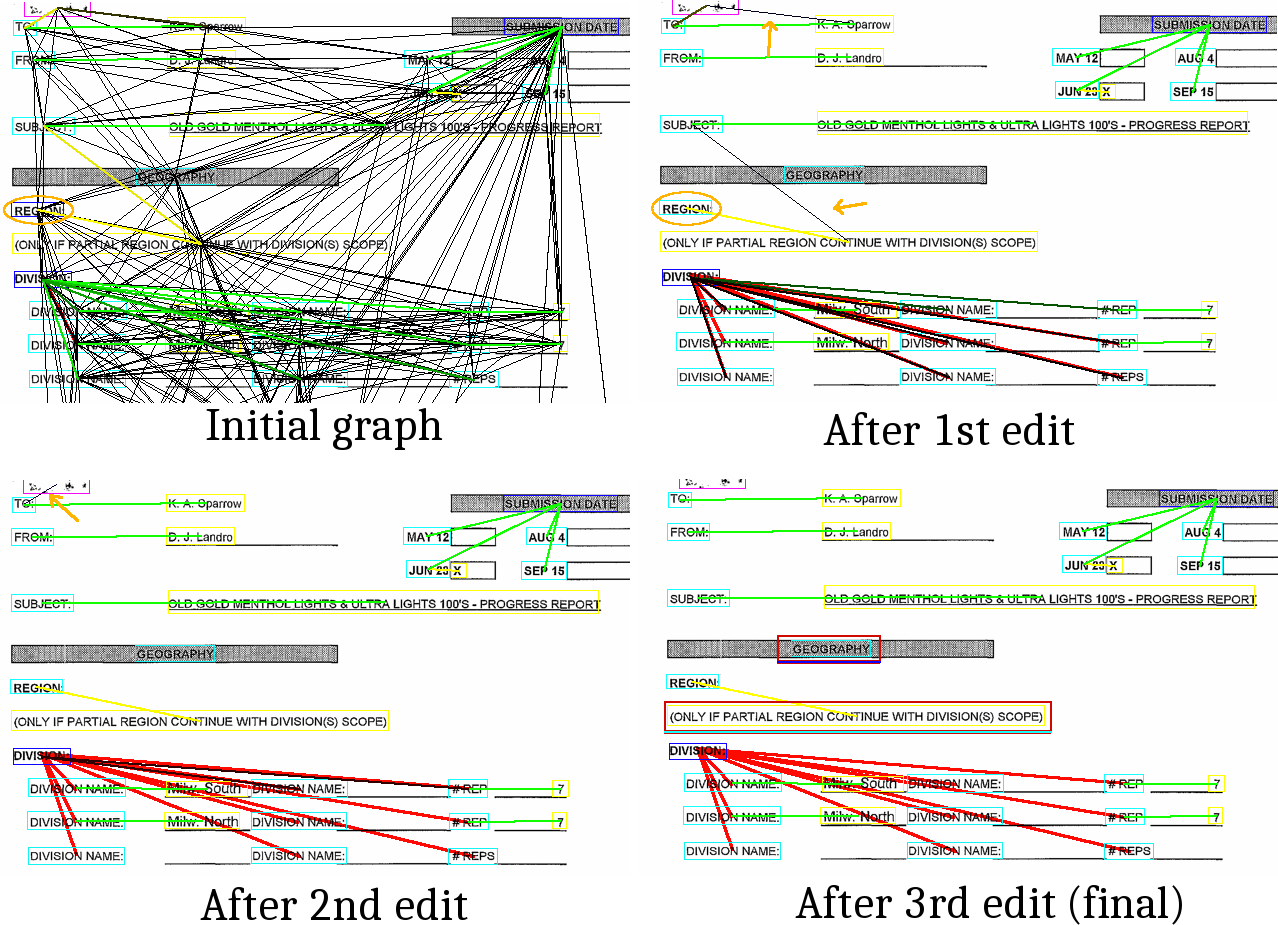}

\caption{Example of iterative graph edits on a FUNSD image. 
Blue, cyan, yellow, and magenta boxes indicate predicted header, question, answer, and other entities. Red boxes are missed entities. Green, yellow, and red lines indicate true-positive, false-positive, and false-negative relationship predictions. Orange marks draw attention to edits.}
\label{fig:edits}
\end{figure}

We use a series of three GCNs, each performing the same iterated predictions. We apply the first GCN to the initial graph and then use its predictions to update the graph structure and features. The next GCN is then applied to the updated graph, which is updated again, and so forth. The final graph contains the (predicted) text entities and relationships.
This process is seen both at the bottom of Fig.~\ref{fig:overview}  and as actual predictions in Fig.~\ref{fig:edits}.

Each GCN in composed of several \textit{GN blocks}, as outlined in Sec.~3.2 of Battaglia et al.~\cite{meta}, without the global attributes and using attention to aggregate edge features for the node update. The GN block first updates the edge features from their two node features and then updates node features from their edge features. The GN block is directed, so we duplicate edge features to create edges in both directions. At the end of a GCN, the predictions and features for the two directions are averaged for each edge. 

Our \textbf{GN block edge update} concatenates the current edge features with its two nodes' features. These are passed to a two-layer fully connected  ReLU network. 
The output is summed with the previous features (residual) to produce the new edge features.

Our \textbf{GN block edge aggregation} (for a given node) applies multi-head attention~\cite{transformer} using the node's features as the query and its edges' features as the keys and values.  
We use 4 heads.

Our \textbf{GN block node update} first appends the aggregation of the edge features  with the node's current features. This is passed through a two-layer fully connected ReLU network. The output is summed with the previous features (residual) to produce the new node features.

All the GCNs use an input and hidden size of 256 (all linear layers are 256 to 256). The first two GCNs have 7 layers, the last has 4; this is based on the observation that most decisions are made earlier on, and so more power is given to the earlier GCNs. 

Each node of the GCN predicts the semantic class for that text entity (or incomplete entity).
Each edge predicts four things: (1) whether the edge should be \textit{pruned}, (2) whether the entities should be \textit{grouped} into a single entity, (3) if these are oversegmented text lines and should be \textit{merged} (corrected), and (4) if this is a true \textit{relationship}.
We threshold each of these predictions and update the graph accordingly. The final graph update uses the relationship prediction to prune edges (so remaining edges are relationships).


If two or more nodes are to be grouped or merged, their features are averaged and their edges are aggregated. Any resulting duplicate edges have their features averaged.
If two nodes are to be merged to fix an oversegmentation, the bounding box for the text line is replaced by the one encompassing both, and their class predictions are averaged.

The features introduced with the initial graph are based on the original bounding boxes, which are potentially modified during a group or merge edit. We reintroduce the initial features again at the start of each GCN. We reuse the same initial features (before the transition layer) for unmodified nodes and edges, and compute new ones for the modified nodes and their edges. These are appended to the final features of the previous GCN and passed through a single linear node or edge transition layer to become the features of the new graph (each GCN has its own transition layers).


As an ablation, we also show results of FUDGE without a GCN, 
where the GCN is replaced by two fully-connected networks which predict from the initial edge and node features respectively.
The graph is still updated in the same way, it is just the GCN being replaced by individual fully-connected networks. To compensate for the lack of complexity, the non-GCN network's fully-connected networks have double the number of hidden channels as the GCN.

The thresholds used to determine if an edit will be made are different on each iteration and were heuristically chosen. The specific thresholds are:
1st edit \{merge: 0.8, group: 0.95, prune: 0.9\},
2nd edit \{merge: 0.9, group: 0.9, prune: 0.8\},
3rd edit \{merge: 0.9, group: 0.6, prune: 0.5\}.
Merge thresholds are initially lower, since we want merges to occur first. Grouping is a higher level decision and so its threshold is higher initially. The prune decision is kept relatively high until the final edit as it's generally desirable to keep edges around. 

We use GroupNorm~\cite{groupnorm} and Dropout in all fully connected and convolutional networks.







\section{Training}
\label{sec:train}

We train the detector first and then train the other components while continuing to fine-tune the detector.
The detection losses are based on YOLO~\cite{yolov3} and are identical to~\cite{previous}. The edge proposals, the GCN edge predictions and the GCN node (class) predictions are all supervised by binary cross-entropy losses. 

When computing the GCN losses, we align the predicted graph to the ground truth (GT) by assigning each predicted text line to a GT text line. From these the proper edge GT can be determined.
We assign predicted and GT text lines by thresholding (at 0.4) a modified IOU which optimally clips the GT bounding boxes horizontally to align them with the predictions; this allows the correct assignment of oversegmented predictions.
If multiple text line predictions are assigned to the same GT text line bounding box, the edges between their nodes them are given a merge GT.
Any edges between nodes with predicted text lines assigned to GT bounding boxes that are part of the same GT text entity are given a grouping GT.
Any edges between two nodes with text line predictions that are assigned to GT text lines which are part of two GT entities with a GT relationship between them are given a relationship GT.
Any edges which don't have either a merge, group, or relationship GT are given GT to be pruned.
For the edge proposal GT, the prune GT is computed for all possible edges.
Nodes are given the GT class of their assigned GT text entities.

Because of memory restrictions we cannot train on an entire form image. Instead, we sample a window of size 800 $\times$ 800 for the FUNSD dataset and 600 $\times$ 1400 for the NAF dataset. We use a batch size of 1.
We also randomly rescale images as a form of data augmentation. The scale is uniformly sampled from a range of 80\%--120\% and preserves aspect ratios.

We train using the AdamW optimizer~\cite{adamw} with a weight decay of 0.01. The detection-only pre-training uses a learning rate of 0.01, and increases the learning rate from zero for the first 1000 iterations, training a total of 250,000 iterations. The full training uses a learning rate of 0.001, increasing the learning rate from zero for the first 1000 iterations. The detector's pre-trained weights are frozen for the first 2000 iterations. In half of the iterations (randomly assigned) during training, we create the initial graph using GT text line bounding boxes. At 590,000 iterations, we drop the learning range by a factor of 0.1 over 10,000 iterations. We then apply stochastic weight averaging (SWA)~\cite{swa} for an additional 100,000 iterations, averaging at every iteration.


\section {Datasets}
Form understanding is a growing area of research, but there are only limited results on public datasets available with which to compare. 
There are two large public datasets of form images annotated for our relationship detection task: the FUNSD dataset~\cite{funsd} and the NAF dataset~\cite{previous}. We did all development on the training and validation sets only.

The \textbf{FUNSD dataset}~\cite{funsd} contains 199 low resolution scans of modern forms. The FUNSD dataset has 50 images as a test set; we divide the training images into a 120 image training set and a 19 image validation set. The forms mostly contain printed text, though some handwriting is present. The images of the FUNSD dataset are relatively clean, though low resolution. The FUNSD dataset is labeled with word bounding boxes with the corresponding transcription, grouping of words into semantic or text entities (one or more lines of text) with a label (header, question, answer, or other), and relations between the text entities. We preprocess the data to group the words into text lines.

The \textbf{NAF dataset}~\cite{previous} contains images of historical forms, 77 test set images, 75 validation set images, and 708 training set images. The images are high resolution, but the documents have a good deal of noise. Most of the forms are filled in by hand and some forms are entirely handwritten. The dataset is labeled with text line bounding boxes with two labels (preprinted text, input text) and the relationships between the text lines. Unlike the FUNSD dataset, there isn't a notion of text entities, rather all the lines which would compose a text entity merely have relationships connecting them. Additionally, the text transcription is unavailable for the NAF dataset, meaning it cannot be used by methods that rely on language. We resize these images to 0.52 their original size.

In Davis et al.~\cite{previous} we only evaluated on a subset of the NAF dataset, the forms which do not contain tables or fill-in-the-blank prose (e.g.~``I \rule{0.7cm}{0.15mm}, on the \rule{0.3cm}{0.15mm} day of \rule{0.7cm}{0.15mm}, do hereby...''). In this work we use the full dataset, although we ignore tables; tables are not annotated and the models learn to ignore them. In \cite{previous}~we only detected relationships between preprinted and input text (key-value), not relationships between text lines of the same semantic class (which would indicate being the same text entities). 
Here we evaluate using all relationships. 

\begin{figure}[]
\includegraphics[width=0.999\textwidth]{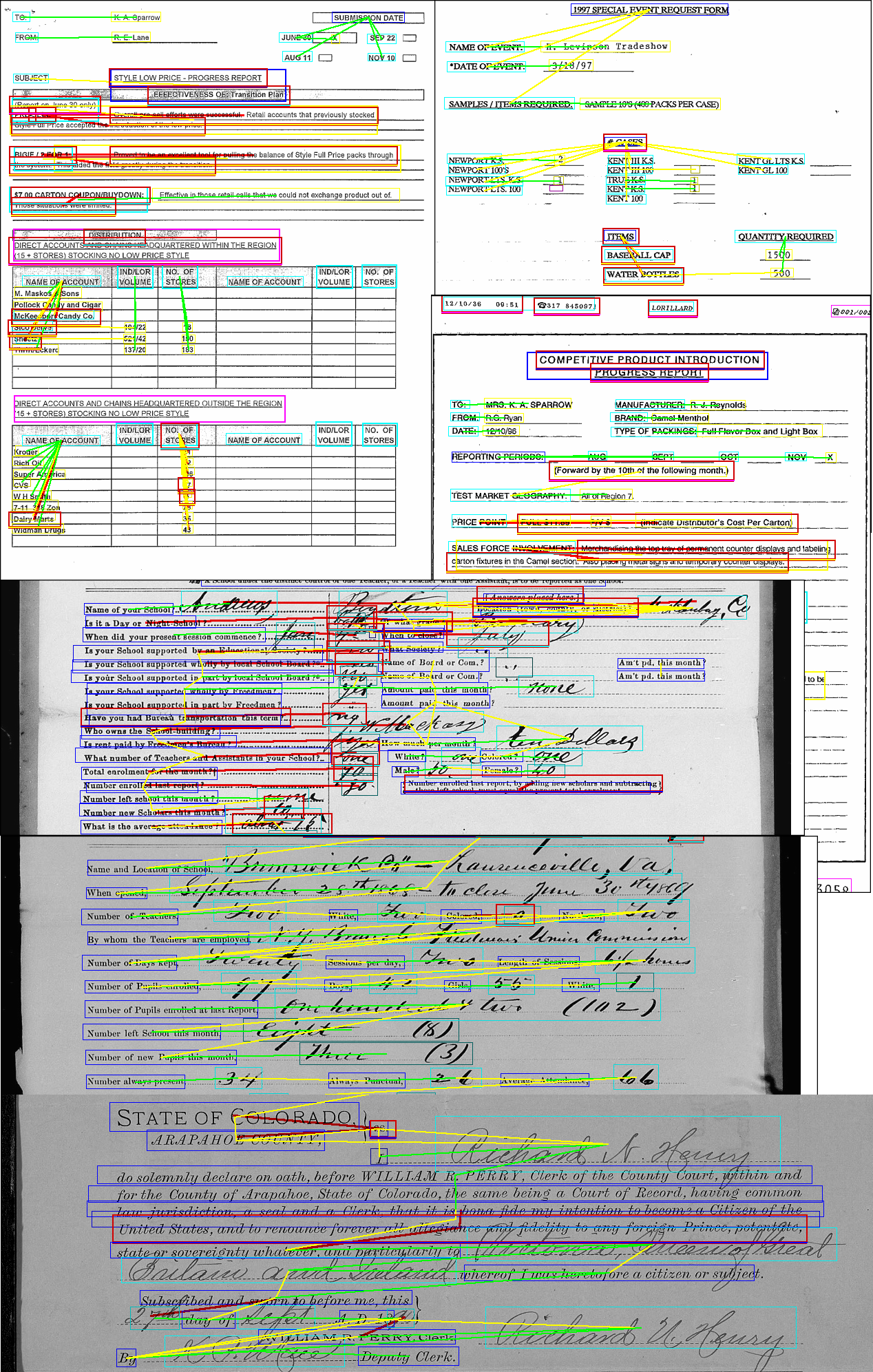}

\caption{Example results of FUDGE's final decisions on crops of three FUNSD images and three NAF images. Blue, cyan, yellow, and magenta boxes indicate predicted header, question, answer, and other entities for the FUNSD images. Red boxes indicate a missed entity (correct class is colored on bottom of red box). Green, yellow, and red lines indicate true-positive, false-positive, and false-negative relationship predictions.}
\label{fig:qual}
\end{figure}

\section{Evaluation and Results}

Qualitative results can be seen in Fig.~\ref{fig:qual}.
Quantitatively, we evaluate performance using the relationship and entity detection (micro) recall, precision, and F1 score (F-measure).
For a correct text entity detection, the prediction must have a bounding box overlapping (with at least 0.5 IOU) each of the text lines making up a ground truth text entity, contain no additional text lines, and have the correct semantic label (class). We adopt the method of scoring of relationships from~\cite{bros}; for a predicted relationship to be correct, the two predicted entities it is between must contain at least the first text line (word in~\cite{bros}) and correct class of two ground truth entities with a relationship. For the NAF dataset, each text line is its own text entity.
In  Davis et al.~\cite{previous} we introduced the NAF dataset, but only evaluated on a simplified subset of the data and only on key-value relationships. In this work we use all the images and relationships of the NAF dataset.
We retrain~\cite{previous} on this larger and harder dataset.
We train~\cite{previous} for 600,000 iterations and use SWA for 100,000 iterations just as we trained FUDGE. We previously \cite{previous} used far fewer iterations as it as on a much smaller dataset. The inclusion of SWA to \cite{previous} makes the comparison more fair as SWA provides a significant boost to performance.
We also train FUDGE on the simplified NAF dataset.
The relationship detection accuracy for both the full and simple subset of the NAF dataset are reported in Table~\ref{tab:all_naf}.  We also report the text detection accuracy in Table~\ref{tab:detect_naf} for the full NAF. 
As can be seen in Table~\ref{tab:all_naf}, FUDGE significantly outperforms \cite{previous} in relationship detection for both the simplified and full NAF dataset. FUDGE performs similarly to \cite{previous} on text line detection, which is reasonable as they share the same text detection backbone. 

\begin{table}[t]
\centering
\caption{Relationship detection (entity linking) on the NAF dataset}\label{tab:all_naf}
\begin{tabular}{|l|r|r|r|}
\multicolumn{4}{l}{Full NAF, all relationships}\\
\hline
Method &  Recall & Precision & F1\\
\hline
Davis et al.~\cite{previous} & 54.62&	45.53&	49.60 \\
FUDGE no GCN & 61.23&	51.96&	56.21 \\ 
FUDGE & 59.92	&54.92	&\textbf{57.31} \\

\hline

\multicolumn{4}{l}{}\\
\multicolumn{4}{l}{Simple subset of NAF, key-value relationships only}\\
\multicolumn{4}{l}{(averaged per document as in \cite{previous})}\\
\hline
Method &  Recall & Precision & F1\\
\hline
Davis et al.~\cite{previous} & 59.9 & 65.4 & 60.7 \\
FUDGE  & 63.6 & 73.2 & \textbf{66.0} \\
 \hline
\end{tabular}
\end{table}

\begin{table}[t]
\centering
\caption{Text line detection on the full NAF dataset}\label{tab:detect_naf}
\begin{tabular}{|l|r|r|r|}
\hline
Method &  Recall & Precision & F1 \\
\hline
Davis et al.~\cite{previous} & 83.23&	66.20&	73.75 \\
FUDGE no GCN & 81.34&	69.45&	74.93  \\
FUDGE & 80.16&	68.22&	73.71 \\
\hline
\end{tabular}
\end{table}

\begin{table}[t]
\centering
\caption{Relationship detection (entity linking) on FUNSD dataset}\label{tab:rel_funsd}
\begin{tabular}{|l|l|l|l|l|l|}
\hline
Method & GT OCR info & \# params&  Recall & Precision & F1  \\
\hline
Carbonell et al.~\cite{carbonell} & word boxes + transcription & 201M& - &-& 39\\
LayoutLM (reimpl.)~\cite{bros}& word boxes + transcription & -& 41.29 & 44.45& 42.81 \\
BROS~\cite{bros} & word boxes + transcription & 138M& 64.30 &69.86& \textbf{66.96}\\

Word-FUDGE&  word boxes  & 17M& 58.08&	67.83&	62.58\\ 
FUDGE no GCN & none & 12M& 50.81	&54.18	&52.41 \\
FUDGE & none & 17M& 54.04	&59.49	&56.62\\
\hline

\end{tabular}
\end{table}




\begin{table}[t]
\centering
\caption{Text entity detection/Semantic labeling on the FUNSD dataset}\label{tab:detect_funsd}
\begin{tabular}{|l|l|l|l|l|l|}
\hline
Method &  GT OCR info &\# params & Recall & Precision & F1  \\
\hline
Carbonell et al.~\cite{carbonell} & word boxes + transcription & 201M& - &-& 64\\
LayoutLM\textsubscript{BASE}~\cite{layoutlm}& word boxes + transcription& 113M & 75.97 & 81.55& 78.66\\
LayoutLM\textsubscript{LARGE}~\cite{layoutlm}& word boxes + transcription&343M & 75.96 &82.19& 78.95 \\
BROS~\cite{bros}& word boxes + transcription& 138M &80.56& 81.88 &81.21  \\
LayoutLMv2\textsubscript{BASE}~\cite{layoutlmv2}& word boxes + transcription& 200M & 80.29 &85.39& 82.76\\
LayoutLMv2\textsubscript{LARGE}~\cite{layoutlmv2}& word boxes + transcription& 426M & 83.24 &85.19 &\textbf{84.20}\\

Word-FUDGE&  word boxes  & 17M& 69.37	&75.30	&72.21\\ 
FUDGE no GCN & none & 12M& 63.64	&66.57	&65.07 \\
FUDGE & none & 17M & 64.90	&68.23	&66.52\\

\hline
\end{tabular}
\end{table}


\begin{table}[t]
\centering
\caption{Hit@1 on FUNSD dataset with GT text entities}\label{tab:withdocstruct} 
\begin{tabular}{|l|l|l|}
\hline
Method &  Hit@1\\
\hline
DocStruct without visual features~\cite{wang-etal-2020-docstruct} &  55.94 \\
DocStruct~\cite{wang-etal-2020-docstruct} &  58.19 \\
FUDGE no GCN & 66.48 \\
FUDGE &\textbf{68.28} \\
\hline
\end{tabular}
\end{table}

In Table~\ref{tab:rel_funsd} we report the same relationship detection metrics for the FUNSD dataset. We compare against  Carbonell et al.~\cite{carbonell} and the unpublished BROS~\cite{bros}  method. 
Davis et al.~\cite{previous} cannot be directly applied to the FUNSD dataset as it lacks a method of grouping the text lines into text entities. 
Carbonell et al. and BROS both use the dataset provided OCR word bounding boxes and transcriptions. 
To compare to these we train Word-FUDGE, which is FUDGE trained using the provided OCR word boxes instead of the line boxes when initializing the graph with the ground truth detections. Word-FUDGE sees the word boxes as oversegmented lines and learns to merge them into text lines.
For relationship detection, Word-FUDGE almost performs as well as BROS, a method that is pre-trained on over ten million additional document images before being fine-tuned on the FUNSD training set, whereas we use \textit{only} the FUNSD training set. 
We think this shows that while the relationship detection problem can be solved with a language-centered approach like BROS, it requires far more data than a visual approach to reach the same performance. We expect a superior approach would combine strong visual features with pre-trained language information. 
The non-GCN version of FUDGE performs almost as well as the GCN version (see also Table~\ref{tab:all_naf}), indicating that predicting in context is either not very necessary or that FUDGE is unable to learn to use the GCN effectively.

Xu et al.~\cite{layoutlm} first presented a semantic labeling task for the FUNSD dataset, which is to predict the semantic entities with their labels, given the word bounding boxes and their transcription. 
We compare our text entity detection against various other methods in Table~\ref{tab:detect_funsd}, both normally and using ground truth word bounding boxes.
Our model is outperformed in this metric by the language-centered approaches. This is understandable; while understanding the layout and having vision helps for this task, it isn't as essential if the language is understood well enough. FUDGE with and without the GCN perform the same, which is surprising as we would expect the context provided by the GCN would improve class predictions.

DocStruct~\cite{wang-etal-2020-docstruct} 
evaluated the relationship detection problem as a retrieval problem, where a query child must retrieve its parent (answers retrieve questions, questions retrieve headers). This view of the problem doesn't account for predicting which nodes have parents in the first place. 
We compare results on the Hit@1 metric, a measure of how often, for each child query, the parent is correctly returned as the most confident result. DocStruct~\cite{wang-etal-2020-docstruct} uses the ground truth OCR text boxes and transcriptions, and also uses the ground truth grouping of text entities (``text fragment'' in~\cite{wang-etal-2020-docstruct}). We also use this information for comparison 
by forcing FUDGE to make the correct text grouping in its first graph edit step, and preventing any further grouping. Our results are compared in Table~\ref{tab:withdocstruct}. 
FUDGE significantly outperforms DocStruct~\cite{wang-etal-2020-docstruct} with and without the GCN. 
However, we don't feel this metric demonstrates general performance as it relies on ground truth text entity grouping and does not measure the ability to detect if a relationship does not exist for a query.

While the previous results have validated the use of the GCN for relationship detection, we also perform an ablation experiment exploring other aspects of FUDGE: the number of graph edit steps, our edge proposal network compared to the line-of-sight proposal used in~\cite{previous}, and the impact of being able to correct text line detections. 
The results are presented in Table~\ref{tab:ablate}. 
Using graph editing and our improved edge proposal improve overall performance. In particular, having at least one intermediate edit step improves entity detection. 
The merging of oversegmented text lines makes only a little improvement, which is to be expected given that the text line detection makes few errors on the FUNSD dataset.
Because we did not perform an exhaustive hyper-parameter search for our primary model, some different choices in the number of GCN layers leads to slightly better results than our primary model.
\begin{table}[t]
\centering
\caption{Ablation models' relationship and entity detection on FUNSD dataset}\label{tab:ablate}
\begin{tabular}{|l|l|l|l|r|r|r|r|r|r|}
\hline
 GCN &    Edit&    Edge &    Allow& \multicolumn{3}{c|}{Relationship detection}& \multicolumn{3}{c|}{Entity detection}\\
 layers &  iters & proposal &  merges &  Recall & Precision & F1 & Recall & Precision & F1\\
\hline

10 & 1 & network & Yes & 59.98&	45.84&	51.96&	66.58&	63.02&	64.75\\
8,8 & 2 & network & Yes & 56.45&	59.22&	\textbf{57.80}&	62.58&	66.35&	64.41\\
3,3,3 & 3 & network & Yes & 54.85&	59.50&	57.08&	65.44&	69.14&	\textbf{67.24}\\
3,3,3 & 3 & line-of-sight & Yes & 48.14&	63.89&	54.90&	65.39&	68.92&	67.11\\
3,3,3 & 3 & network & No & 55.73&	57.97&	56.83&	66.99&	68.10&	\textbf{67.44}\\
7,7,4 & 3 & network & Yes & 54.04	&59.49	&56.62 & 64.90	&68.23	&66.52\\
\hline
\end{tabular}
\end{table}



Looking at the relationship detection errors made by FUDGE in detail, it can be seen that the majority of false negative relationships are actually caused by poor entity detection, which is why the performance increases so dramatically with the use of ground truth bounding boxes.

\section{Conclusion}

We present FUDGE, a visual approach to form understanding that uses a predicted form graph and edits it to reflect the true structure of the document. The graph is created from detected text lines (vertices) and the initial edges are proposed by a simple network. In three iterations, GCNs predict whether to combine vertices to group text elements into single entities or to prune edges.

FUDGE uses no language information, but it performs similarly to methods using large pre-trained language models. 
We believe this demonstrates that most form understanding solutions do not put enough emphasis on visual information.
%



%
%
%
 \bibliographystyle{splncs04}
 \bibliography{bib}
\end{document}